\documentclass{article} 
\usepackage{nips14submit_e,times}
\usepackage{url}
\usepackage{graphicx}
\usepackage[sort&compress,numbers]{natbib}

\title{Using Sentence Plausibility to Learn the Semantics of Transitive Verbs}

\author{
Tamara Polajnar~~~~~Laura Rimell~~~~~Stephen Clark\\
Computer Laboratory\\
University of Cambridge\\
United Kingdom\\
\texttt{\{tamara.polajnar,laura.rimell,stephen.clark\}@cl.cam.ac.uk} \\
}

\nipsfinalcopy 

\begin{document}

\maketitle

\begin{abstract}
The functional approach to compositional distributional semantics
considers transitive verbs to be linear maps that transform the
distributional vectors representing nouns into a vector representing a
sentence. We conduct an initial investigation that uses a matrix consisting of the
parameters of a logistic regression classifier trained on a
plausibility task as a transitive verb function. We compare our method to a
commonly used corpus-based method for constructing a verb matrix and
find that the plausibility training may be more effective for
disambiguation tasks.
\end{abstract}

\section{Introduction}

The field of compositional distributional semantics seeks principled
ways to combine distributional representations of words to form larger
units. Representations of full sentences, besides their theoretical
interest, have the potential to be useful for tasks such as automatic
summarisation and recognising textual entailment. A number of recent
studies have investigated ways to combine distributional
representations of Subject, Verb, Object (SVO) triples to form
transitive sentences, that is, sentences based on a transitive verb
\cite{Mitchell:Lapata:08,BaroniEMNLP10,LambekFest,gref:emnlp11,clark:13,Grefenstette13,polajnar:emnlp14}.

Under the {\it functional} approach
\cite{BaroniEMNLP10,LambekFest,baroni:frege,clark:13}, argument-taking words
such as verbs and adjectives are represented as tensors, which take
as arguments word representations of lower order, typically vectors. A transitive verb can be viewed as a
third-order tensor with input dimensions for the subject and object,
and an output dimension for the meaning of the sentence as a
whole. This approach has achieved promising initial results
\cite{gref:emnlp11,gref:gems11,Grefenstette13,kartsaklis:emnlp13,polajnar:emnlp14},
but many questions remain. Two outstanding questions are the most effective 
method of learning verb tensors from a corpus, and the most appropriate 
sentence space for a variety of different tasks.

This paper presents work in progress which addresses both of these
questions. It compares three methods for learning verb representations:
the distributional model of \cite{gref:emnlp11} in which positive
examples of subject-object pairs for a given verb are structurally
mixed, the regression model of \cite{polajnar:emnlp14} in which
positive and negative examples of subject-object pairs for a given verb are mapped into
a plausibility space, and a variant of the regression model which uses only positive examples. A variety of methods for composing the verb, subject, and object representations are
investigated. The results show that the plausibility training
outperforms the distributional method on a verb disambiguation task,
while the purely distributional approach performs better on sentence
similarity.

\section{Methods}


In the following experiments we consider a transitive verb as a map that takes
noun vectors representing the subject and object as arguments, and produces a
vector in the sentence space. Typically, noun vectors for subject and
object reside in a ``context space'' where the dimensions correspond to
co-occurrence features; we use a reduced space resulting from applying
Singular Value Decomposition (SVD) to the co-occurrence space. The most appropriate sentence space to use
is less obvious; previous approaches have either mapped sentence
meaning to the same topic-based noun space
\cite{gref:emnlp11,Grefenstette13} or defined a new space for sentence
meaning, particularly plausibility space
\cite{Krishnamurthy13,polajnar:emnlp14}.

If the verb function is a multi-linear map, then the verb is naturally
represented by a third-order tensor. However, tensor training can be
expensive and in practice, for some tasks, the verb can be approximated
as a matrix \cite{gref:emnlp11,polajnar:emnlp14}.

Below we describe three ways of learning a verb matrix.  In the
distributional method, training is based on a sum of plausible
(i.e. attested) subject-object pairs for a particular verb. In the
regression method, the learnt matrix consists of parameters from a
plausibility classifier. The classifier is trained to distinguish
plausible sentences like {\em animals eat plants} from implausible
sentences like {\em animals eat planets}.  In the regression-positive
method, only plausible training examples are used. The acquisition of
training data for all three methods 
is described in
Section~\ref{trdata}.

\subsection{Verbs}

\paragraph{Distributional (dist)} Following \cite{gref:emnlp11},  
we generate a $K\times K$ matrix for each verb as the average of outer products ($\otimes$) of $K$-dimensional subject and object vectors 
from the positively labelled subset of the training data:


{\small
$$
\mathrm{V} = \frac{1}{N_p} \left[\sum_{i=1}^{N_p} \overrightarrow{s}_i \otimes \overrightarrow{o}_i\right]
$$
}


where $N_p$ is the number of positive training examples. \noindent The intuition is
that
the matrix encodes higher weights for contextual features of frequently attested subjects and objects; 
for example, multiplying by the matrix for {\it eat} may yield a higher scalar value when its subject exhibits features common to animate nouns, and 
its object exhibits features common to edible nouns. 

\paragraph{Regression (reg)} 

Following \cite{polajnar:emnlp14}, we formulate regression learning as a plausibility task where the
class membership can be estimated with a single variable.  To produce
a scalar output, we can learn the parameters for a single $K\times K$
matrix ($\mathrm{V}$) using standard logistic regression with the mean
squared error cost function and $K$-dimensional subject ($\overrightarrow{s}$) and
object ($\overrightarrow{o}$) noun vectors as input:


{\small
$$
O(\mathrm{V}) = - \frac{1}{m}\left[\sum_{i=1}^N
t_i\log h_{\mathrm{V}}\left(\overrightarrow{s}_i,\overrightarrow{o}_i\right) +(1-t^i)\log h_{\mathrm{V}}\left(\overrightarrow{s}_i,\overrightarrow{o}_i\right)\right] 
$$
}


\noindent
where $t_i$ are the true plausibility labels (1 or 0) of the $N$ training examples. The function $h_{\mathrm{V}}\left(\overrightarrow{s}_i,\overrightarrow{o}_i\right) = \sigma((\overrightarrow{s}_i)^T\mathrm{V}(\overrightarrow{o}_i))$ is a sigmoid transformation of the scalar that results from the matrix multiplication and the objective is regularised by the parameter $\lambda$:
$O(\mathrm{V}) + \frac{\lambda}{2}||\mathrm{V}||^2$. 

The resulting matrix is the representation of a transitive verb, and although it is trained to produce a point in the plausibility sentence space, we use it in the same manner as {\bf dist}. 
The regression algorithm
is trained through gradient descent with ADADELTA \cite{Zeiler12} and
10\% of the training triples are used as a validation set for early
stopping.

\paragraph{Regression-Positive (reg+)}

We also examine a variant of regression in which we only train with positive examples. The training data is therefore the same as {\bf dist}, but the training method is the same as {\bf reg}.

\subsection{Training data}
\label{trdata}
In order to generate training data we find SVO triples that
occur in an October 2013 dump of Wikipedia. These attested triples are considered plausible data. To ensure quality we
choose triples whose nouns occur at least 100 times. For some verbs
there are thousands of such triples, so we choose the top 1000 most
frequent triples for each verb. For each verb we generate negative
examples by substituting the plausible subject or object or both with
maximally dissimilar nouns.  Specifically, for a given subject (or
object) noun we calculate its average sum with the centroid of
plausible subject (or object) vectors, and then select the
frequency-matched noun with lowest cosine similarity to this
average. For each plausible example we generate three negative
examples, one where both subject and object are substituted, as well
as two examples where the either the subject or object is substituted,
but the other is still correct. We then randomly sample 1000 negative
training points from this pool.
The noun and verb vectors are generated from the Wikipedia corpus using the
t-test weighting scheme and normalisation techniques described in
\cite{Polajnar14}. These techniques enable us to learn high-quality representations using SVD
reduced vectors with dimensions as low as 20.

\subsection{Composition Methods}

We investigate the following methods of composing the verb matrix ($\overline{\mathrm{V}}$) with the subject ($\overrightarrow{s}$) and object ($\overrightarrow{o}$) vectors to form a vector representation for a transitive sentence. We make use of the outer product ($\otimes$), elementwise product ($\odot$), and matrix multiplication ($\times$).

{\bf Copy-object (CO)}: from \cite{kartsaklis:emnlp13,kartsaklis:qpl14}, the meaning of a transitive sentence is a vector, obtained by:

\begin{equation}
\overrightarrow{s\ \mathrm{V}\ o} = (\overrightarrow{s} \times \overline{\mathrm{V}}) \odot \overrightarrow{o} = \{\overrightarrow{s}^T \times \overline{\mathrm{V}}\}_{i} \cdot \overrightarrow{o}_{i}
\end{equation}

{\bf Copy-subject (CS)}: from \cite{kartsaklis:emnlp13,kartsaklis:qpl14}, the meaning of a transitive sentence is a vector, obtained by:

\begin{equation}
\overrightarrow{s\ \mathrm{V}\ o} = \overrightarrow{s} \odot (\overline{\mathrm{V}} \times \overrightarrow{o}) = \overrightarrow{s}_{i} \cdot \{\overline{\mathrm{V}} \times \overrightarrow{o}\}_{i}
\end{equation}

{\bf Frobenius additive (F+)}: from \cite{kartsaklis:qpl14}, the meaning of a transitive sentence is a vector, obtained by addition of the vectors produced by {\bf CS} and {\bf CO}:

\begin{equation}
\overrightarrow{s\ \mathrm{V}\ o} = (\overrightarrow{s} \times \overline{\mathrm{V}}) \odot \overrightarrow{o} + \overrightarrow{s} \odot (\overline{\mathrm{V}} \times \overrightarrow{o})
\end{equation}

{\bf Relational (RE)}: from \cite{gref:emnlp11,kartsaklis:qpl14}, the meaning of a transitive sentence is a matrix, obtained by the following formula:

\begin{equation}
\overline{s\ \mathrm{V}\ o} = (\overrightarrow{s} \otimes \overrightarrow{o}) \odot \overline{\mathrm{V}} = \{\overrightarrow{s}\times \overrightarrow{o}^T\}_{ij}\cdot \overline{\mathrm{V}}_{ij}
\end{equation}

{\bf Verb-object (VO)}: this method tests whether the verb and object alone are enough to measure sentence similarity. It can be compared directly to {\bf CS} and reflects the linguistic generalisation that a verb selects its object more strongly than its subject.
The meaning of a transitive sentence is approximated by a vector encoding the verb phrase:

\begin{equation}
\overrightarrow{s\ V\ o} \approx \overline{\mathrm{V}} \times \overrightarrow{o}
\end{equation}

We also report the following two standard methods that do not take into account argument order, and which use a distributional vector representation of the verb {$\overrightarrow{v}$.

{\bf Additive (Add) }: $ \overrightarrow{s\ v\ o} = \overrightarrow{s} + \overrightarrow{v} + \overrightarrow{o} $

{\bf Multiplicative (Mult) }: $ \overrightarrow{s\ v\ o} = \overrightarrow{s} \odot \overrightarrow{v} \odot \overrightarrow{o} $

For the {\bf Relational} method, sentence similarity is measured as the Frobenius inner product of the two sentence matrices. For the rest of the methods, sentence similarity is measured as the cosine of the two sentence vectors.

\section{Tasks}

We investigate the performance of the regression learning method on two tasks: verb disambiguation, and transitive sentence similarity. In each case the system must compose SVO triples and 
assign similarity values to pairs of composed triples.

For the verb disambiguation task we use the {\bf GS2011} dataset
\cite{gref:emnlp11}. This dataset consists of pairs of SVO triples
in which the subject and object are held constant, and the verb is
manipulated to highlight different word senses. For example, the verb
{\it draw} has senses that correspond to {\it attract} and {\it
  depict}.  The SVO triple {\it report draw attention} has high
similarity to {\it report attract attention}, but low similarity to
{\it report depict attention}. Conversely, {\it child draw picture}
has high similarity to {\it child depict picture}, but low similarity
to {\it child attract picture}. The gold standard consists of human
judgements of the similarity between pairs, and we measure the correlation of the
system's similarity scores to the gold standard judgements.

For the transitive sentence similarity task we use the {\bf KS2013}
dataset \cite{kartsaklis:emnlp13}. This dataset also consists of pairs of
SVO triples, but the subject and object as well as the verb vary. For
example, {\it author write book} and {\it delegate buy land} are
judged by most annotators to be very dissimilar, while {\it programme
  offer support} and {\it service provide help} are considered highly
similar. Again, we measure the correlation between the system's
similarity scores and the gold standard judgements.

\section{Results}

\begin{table}[t!]

\begin{center}
\begin{tabular}{l|l|ll|ll}
\multicolumn{1}{c}{} & \multicolumn{1}{c}{} & \multicolumn{2}{c}{\bf GS2011} & \multicolumn{2}{c}{\bf KS2013}  \\
\multicolumn{1}{c}{\bf METHOD} &\multicolumn{1}{c}{\bf COMP} &
\multicolumn{1}{c}{\bf K=20} & \multicolumn{1}{c}{\bf K=300} & \multicolumn{1}{c}{\bf K=20} & \multicolumn{1}{c}{\bf K=300} \\ \hline 
& & & & & \\
{\bf baseline} & {\bf add} & {\bf 0.17} & 0.12 & 0.48 & {\bf 0.58} \\  
 & {\bf mult} & 0.22 & 0.{\bf 24} & {\bf  0.16} & 0.13 \\ \hline
{\bf dist} & {\bf CO}& 0.26 & 0.30  & 0.19 & 0.24 \\
           & {\bf CS} & 0.23 & 0.28 & 0.24 & 0.24 \\
           & {\bf F+} & {\bf 0.28} & {\bf 0.32} & 0.22 & 0.25 \\
           & {\bf RE} & 0.18 & 0.16 & {\bf 0.27} &  {\bf 0.33}\\
           & {\bf VO} & 0.19 & 0.24 & 0.23 & 0.26\\ \hline

{\bf reg+} & {\bf CO} & 0.26 & 0.30 & 0.20 & 0.27 \\
           & {\bf CS} & 0.24 & 0.29 & 0.23 & 0.23 \\
           & {\bf F+} & {\bf 0.29} & {\bf 0.33} & 0.23 & 0.25 \\
           & {\bf RE} & 0.19 & 0.16& {\bf 0.26} &  {\bf 0.33}\\
           & {\bf VO} & 0.21 & 0.25 & 0.23 & 0.29\\ \hline
{\bf reg} & {\bf CO} & 0.24& 0.33 & 0.19 & 0.27 \\
           & {\bf CS} & 0.23& 0.28 & 0.19 & 0.21 \\
           & {\bf F+} &{\bf 0.28} & {\bf 0.35} & 0.12 & 0.20 \\
           & {\bf RE} & 0.20 & 0.20 & {\bf 0.27} &  {\bf 0.31}\\
           & {\bf VO} & 0.22 & 0.28 & 0.23 & 0.27\\

 \end{tabular}
 \end{center}
\caption{Spearman correlation for each method on {\bf GS2011} and {\bf KS2013} datasets across composition methods and number of noun dimensions. }
\label{restable}
 \end{table}

\begin{figure}[h]
\begin{center}
\hspace{-.1cm}\includegraphics[width=0.48\textwidth]{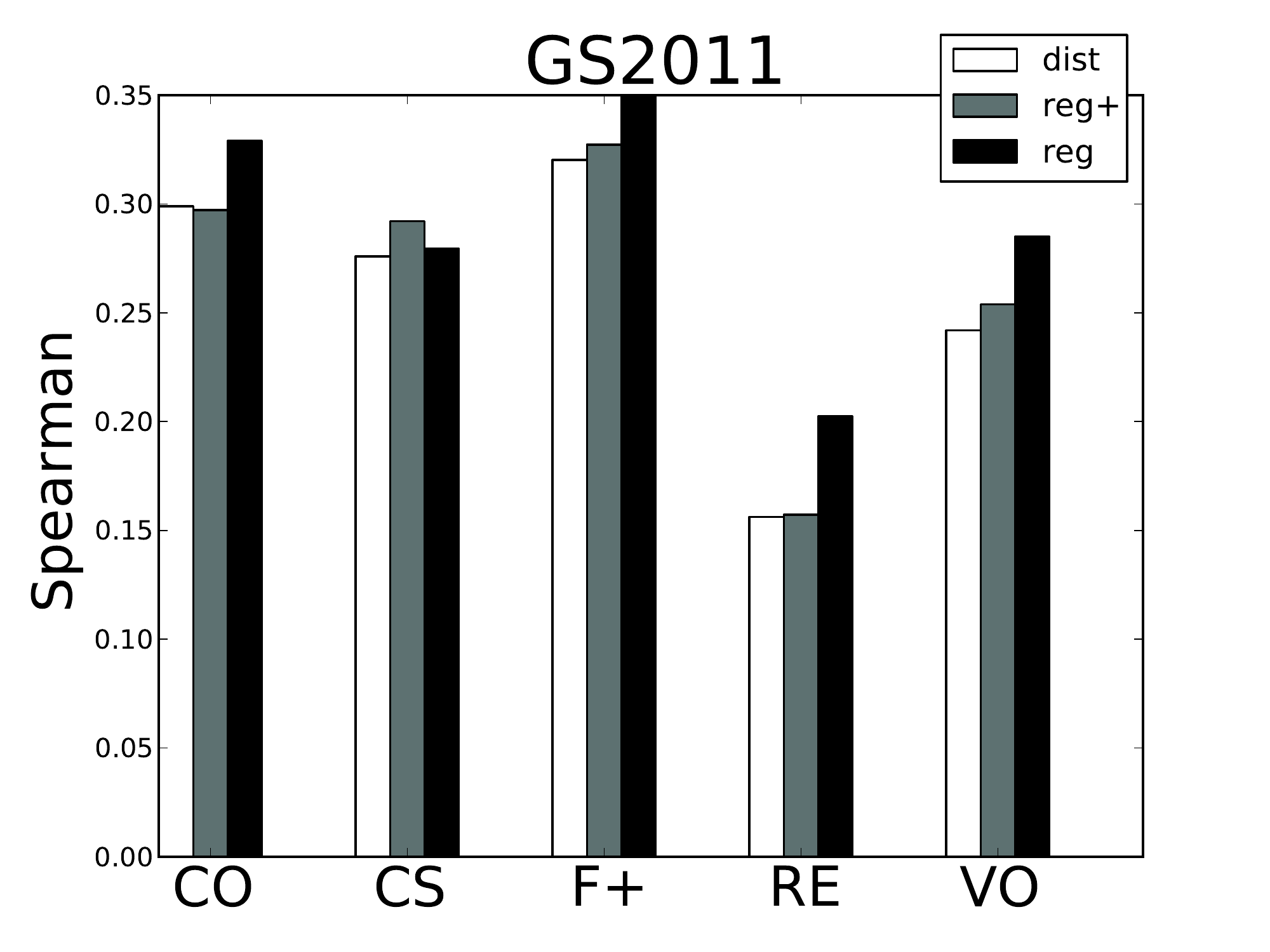}
\hspace{-.1cm}\includegraphics[width=0.48\textwidth]{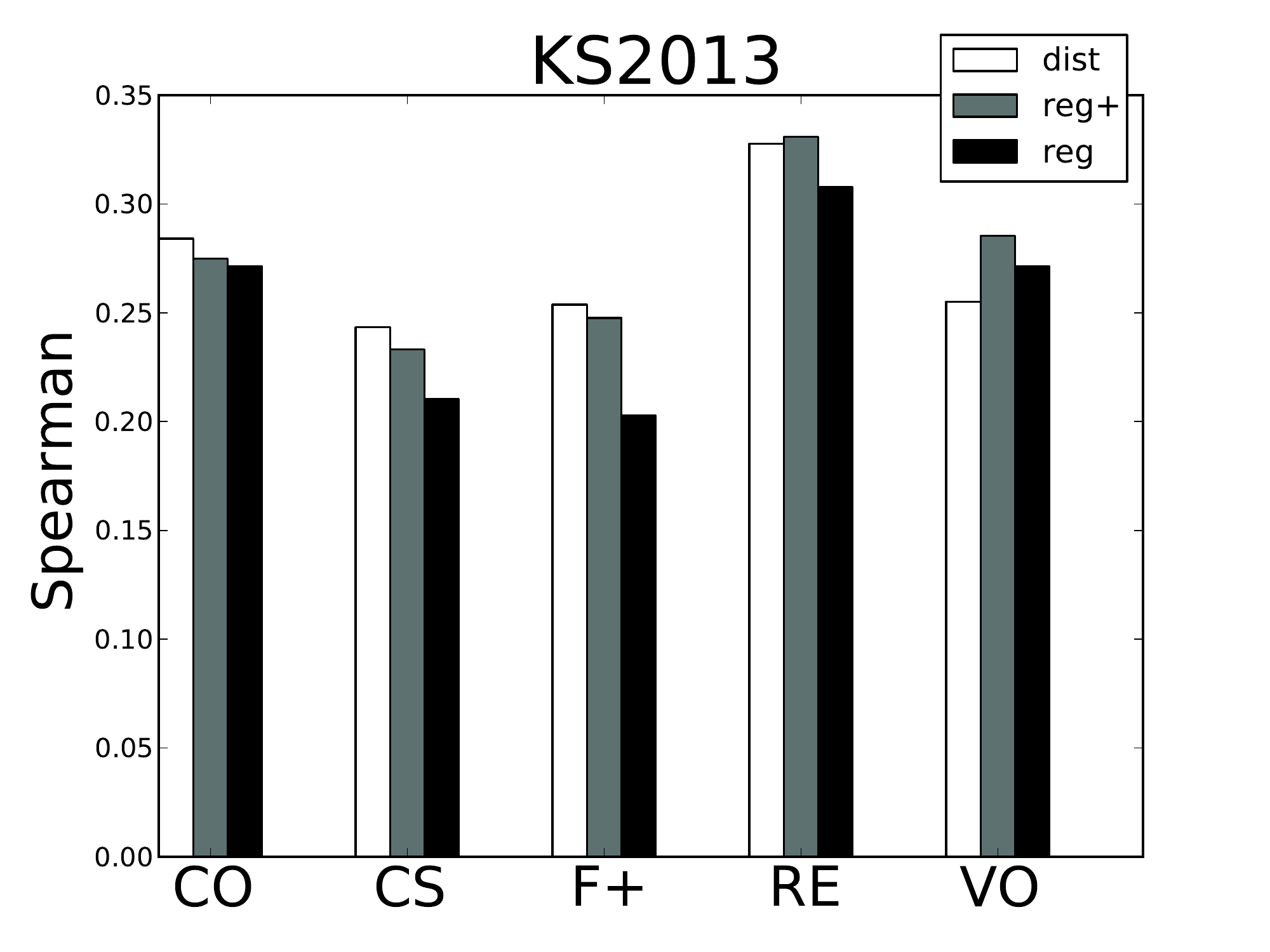}

\end{center}

\caption{Performance of {\bf dist} and {\bf reg} on two datasets {\bf GS2011} and {\bf KS2013} over several composition methods with 300 dimensional vectors.}

\label{resfig}
\end{figure}






Table~\ref{restable} and Figure~\ref{resfig} show the results of
our experiments.  Overall, 
{\bf reg}
performs better than 
{\bf dist}
on verb disambiguation, while 
{\bf dist}
performs better
on sentence similarity. We hypothesise the difference lies in
the nature of the two tasks.  The verb disambiguation task is
inherently plausibility-based, because one member of the
low-similarity pairs (with the non-relevant sense of the
verb) is always implausible. On the other hand, both triples in the sentence
similarity task tend to be highly plausible, even when their topic
differs. Because 
{\bf dist}
uses a topic-based space,
it may better capture these distinctions. 

To investigate this hypothesis, we performed an error analysis using
mean squared error (MSQE). We first averaged the gold labels across
annotators and then we normalised the values from 0~to~1. We also
normalised the output of the cosine similarities between the vectors
produced by the composition methods. Following normalisation, we calculated the MSQE for each pair in order to identify which SVO sentence pairs were furthest away from their average gold standard ranking. We subtracted the per pair MSQE
values of one method from another in order to identify the examples where
methods exhibited divergent behaviour. 

We found that both methods perform well on the highest similarity pairs (those with an average 6 or 7 human rating). On GS2011 using Frobenius additive, the best overall composition method on this dataset, {\bf reg} indeed performed better than {\bf dist} on pairs such as {\it student meet requirement -- student visit requirement}, correctly assigning a low similarity value. Here, the second member of the pair is implausible. We observed no particular pattern to the cases in which {\bf dist} performed better than {\bf reg}.

On KS2013 using Relational composition, the best overall composition method on this dataset, we found that {\bf reg} performed better than {\bf dist} on a number of low-similarity pairs, such as {\it man wave hand -- employee start work}. This seems to run counter to the original hypothesis, namely that {\bf dist} will perform better when the SVO triples in the pair are both plausible, yet dissimilar from each other. However, the training data must also be taken into account. It may be that {\it man wave hand} is an implausible triple in Wikipedia, which causes {\bf reg} to give the pair a lower rating. On this dataset, {\bf dist} performs better than {\bf reg} on a variety of mid-to-high similarity pairs such as {\it project present problem -- programme face difficulty}.

The {\bf reg+} method, which uses the same training data as {\it dist} and training method as {\it reg}, shows an overall pattern similar to {\it dist}, but with higher scores in general on GS2011, and lower with some of the composition methods on KS2013. This is consistent with the idea that the nature of the training data is the important distinction between the plausibility and similarity tasks. It also suggests that using regression can improve the results over the simple distributional method of building matrices.

Table~\ref{restable} also shows that 20-dimensional noun vectors perform with reasonable accuracy compared to 300-dimensional noun vectors. Due to the nature of functional
composition of distributional representations, where each word-type other than noun is
represented by a higher-order tensor, low-dimensional representations
are particularly advantageous. 



\section{Conclusion}

The difference in performance of the two methods underlines the need
to find the appropriate sentence spaces for particular tasks. This
preliminary study indicates that plausibility training may be better
suited for disambiguation. Further work will consist of more in depth
analysis and optimisation of the training procedure, as well as
investigation into ways of low-cost learning of task-specific sentence
spaces.

\subsubsection*{Acknowledgements}

Tamara Polajnar is supported by the ERC Starting Grant, DisCoTex,
awarded to Stephen Clark, and Laura Rimell is supported by the EPSRC
grant EP/I037512/1: A Unified Model of Compositional and
Distributional Semantics: Theory and Applications.

\bibliographystyle{plain}
\bibliography{tp-copy}

\end{document}